\documentclass[10pt,twocolumn,letterpaper]{article}

\usepackage{iccv}
\usepackage{times}
\usepackage{epsfig}
\usepackage{graphicx}
\usepackage{amsmath}
\usepackage{amssymb}
\usepackage[dvipsnames]{xcolor}
\usepackage{caption}
\usepackage{graphicx}
\usepackage{amsmath}
\usepackage{amssymb}
\usepackage{caption}
\usepackage{booktabs}
\usepackage{xcolor,colortbl}

\usepackage[breaklinks=true,bookmarks=false]{hyperref}

\iccvfinalcopy % *** Uncomment this line for the final submission

\def\iccvPaperID{****} % *** Enter the ICCV Paper ID here

\ificcvfinal\fi

\begin{document}

%%%%%%%%% TITLE
\title{Muscles in Action}

\author{Mia Chiquier and Carl Vondrick\\Columbia University\\
{\tt\small \{mia.chiquier,vondrick\}@cs.columbia.edu}
}

\twocolumn[{%
        \renewcommand\twocolumn[1][]{#1}%
        \maketitle
        \begin{center}
        \captionsetup{type=figure}
                \vspace{-0.5cm}
        \includegraphics[width=1.0\linewidth]{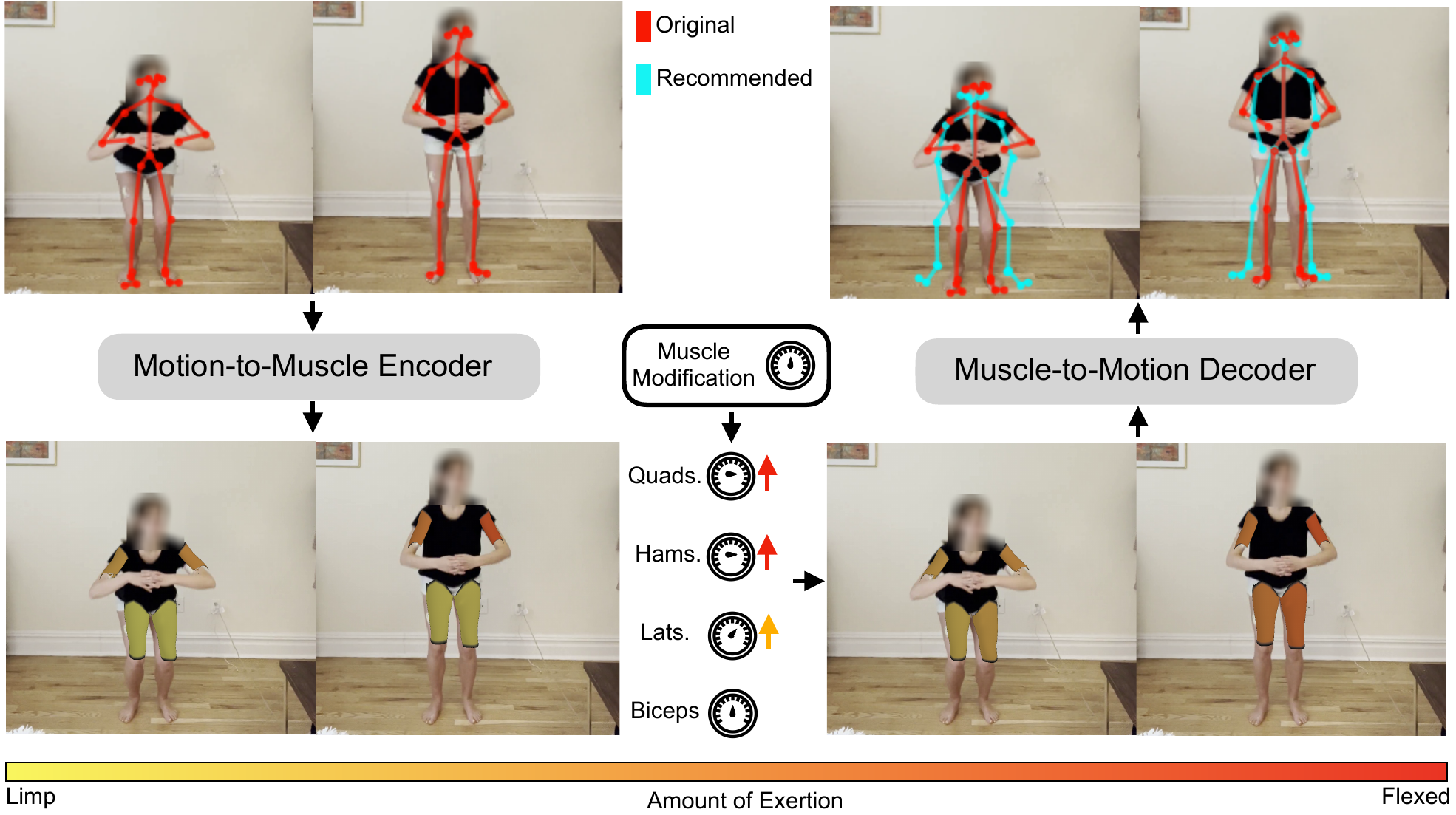}
        \captionof{figure}{\textbf{Bidirectional Mapping between Muscles and Motion}. Visible human motion is created by, and constrained by, our muscles. We learn to predict which muscle groups a person uses during motion (left column), as well as to reconstruct motion from muscle activation (right column). We illustrate how these two mappings can be combined to recommend new motions, similar to the input muscle, this time subject to particular muscle group goals (bottom row).
\vspace{0.1cm}
}
               
                \label{fig:teaser}
        \end{center}
}]
%%%%%%%%% ABSTRACT
\begin{abstract}
\vspace{-1em}
   Human motion is created by, and constrained by, our muscles.
   We take a first step at building computer vision methods that represent the internal muscle activity that causes motion. 
   We present a new dataset,  Muscles in Action (MIA), to learn to incorporate muscle activity into human motion representations. The dataset consists of 12.5 hours of synchronized video and surface electromyography (sEMG) data of 10 subjects performing various exercises. Using this dataset, we learn a bidirectional representation that predicts muscle activation from video, and conversely, reconstructs motion from muscle activation. We evaluate our model on in-distribution subjects and exercises, as well as on out-of-distribution subjects and exercises.
   We demonstrate how advances in modeling both modalities jointly can serve as conditioning for muscularly consistent motion generation.
   Putting muscles into computer vision systems will enable richer models of virtual humans, with applications in sports, fitness, and AR/VR. 
\end{abstract}

%%%%%%%%% BODY TEXT
\vspace{-1em}
\section{Introduction}
\label{sec:intro}
%\textbf{Whats the problem?/What problem does this paper solve?}

% 1.) What problem does this paper solve?
%2.)Why is this important/who is target audience
%3.) What are core technical/conceptual contributions? What can we do now that we cant do before?

% Analyzing human behavior in video is a fundamental problem in computer vision, and many methods have been developed to recognize activities, parse human poses, anticipate their motions, and reconstruct their bodies. Building rich representations of human activities underlies applications in robotics, sports, virtual reality, and health, motivating the need for the field to develop rich representations that learn to capture the structure underlying human motion. However, nearly all methods in computer vision today model only the visible parts of a person, which neglects the rich underlying body structure that physically generates human motion.

The vision community has made great progress in modelling and analyzing human motion from video via tasks such as pose estimation  \cite{loper2015smpl, kocabas2020vibe, yuan2022glamr, pishchulin2016deepcut, cao2017realtime, kay2017kinetics, cao2017realtime}, action recognition \cite{raptis2013poselet,vahdat2011discriminative,kong2017deep,kong2014discriminative,wang2019progressive}, motion transfer \cite{chan2019everybody, aberman2019deep, liu2019neural, wang2018video} and more. However, motion understanding goes beyond the surface. Human motion is created by and constrained by muscles. Every action is a product of our brain sending electric signals to our nerves, which contract our muscles, in turn moving our joints. Although this process occurs within us, most of us turn to physical therapists and sports instructors for guidance on how to improve our motions to target or avoid particular muscle groups. 

In this paper, we take a first step towards building computer vision methods that represent the internal muscle activity that causes human motion. We present a system that, given a video of a person performing an action, learns to infer what muscles a person used. Walking is controlled falling, and any physical motion is a balance between muscle forces and gravity. This interplay leads to an inherent asymmetry: different muscles are engaged in the downward portion of a squat, for instance, than in the upward portion.

Our goal is to learn the complex relationships between physical forces by analyzing synchronized video and muscle activation data. We achieve this by developing a system that can predict muscle activity from motion, and vice versa. One application of this bidirectional system is generating new motions that are similar to an existing motion, while also adhering to specific muscle recruitment targets, illustrated in Figure \ref{fig:teaser}.

% We aim to implicitly learn the rich interactions between these forces by leveraging synchronized video and muscle activation data. We also learn the inverse direction: reconstructing motion from muscle activation. With both learned mappings, our system is bidirectional, allowing us to create a new form of motion generation conditioning: generating motions that are similar to an input motion, with the key difference that the new motion respects specified muscle group recruitment goals. Figure 1 shows one application of our approach, where we are able to provide recommendations to people on how they could modify their physical activity in order to change how much of each muscle they engage, which has applications in sports and physical therapy.

\begin{figure}[t]
\centering
%\framebox[4.0in]{$\;$}
\includegraphics[width=\linewidth]{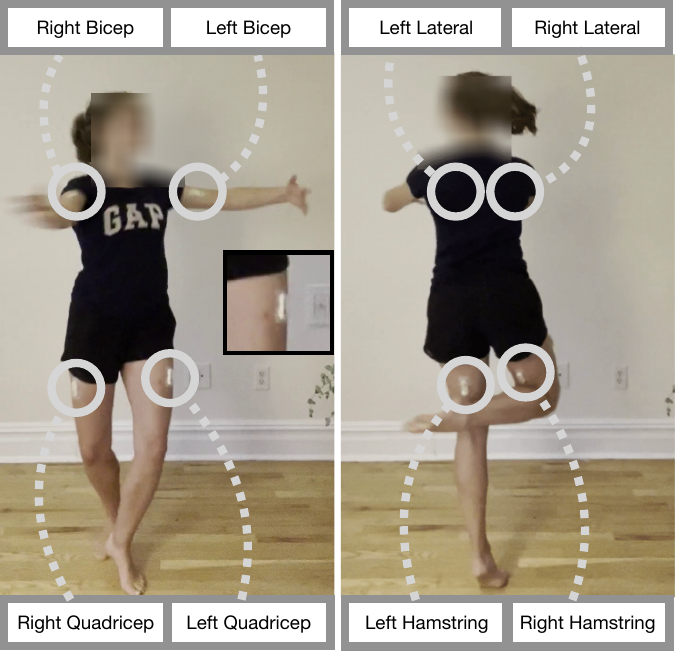}
\caption{\looseness=-1\textbf{Sensor Placement.} We illustrate the placement of our 8 sEMG sensors on a subject. We label the 8 measured muscles.}
 \label{fig:sensors}
 \vspace{-1em}
\end{figure}

\looseness=-1
The typical method of measuring muscle activity is through the use of electromyography sensors, which exist in an invasive form, as well as an non-invasive form, called surface electromyography (sEMG). We collected a new dataset, which we will release, that consists of over twelve hours of synchronized single-view video and sEMG signals of eight muscles for ten subjects performing fifteen different physical activities. By using commodity cameras and inexpensive sEMG sensors, we make the problem practical and easy for others to build on. The eight muscles recorded are the left and right biceps brachii (biceps), the left and right latissimus dorsi (laterals), both quadriceps (quads), and both biceps femoris (hamstrings), denoted in Figure \ref{fig:sensors}.  

% By learning from the multi-modal synchronization between the visual modality and the muscle modality, experiments show that our method is able to recover which muscle groups are used for different activities from video, as well as reconstruct motion from muscle activation. 

%In addition to evaluating our models on in-distribution subjects and exercises, we also evaluate on out-of-distribution subjects and experiments. Our analysis of these experiments convey two important insights: a) our learning based methods generalize better than non-learning based baselines, b) out-of-distribution exercises that have more visually similar exercises in the in-distribution set generalize better than those that have fewer. The first point serves as a benchmark for future work, and the second point validates our hypothesis that the key to success lies in a uniformly sampled dataset. In other words, the result serves to guide which exercises to add as we, and the community, contribute further to the MIA dataset. 

The primary contribution of this paper is a framework for modeling the association between human motion and internal muscle activity in video, and the rest of the paper will explain this contribution in detail. In section 2, we briefly review related work in human activity analysis, conditional motion generation, multi-modal learning, electromyography, and physics-grounded human motion generation. In section 3, we describe our multimodal dataset in detail and analyze its characteristics. In section 4, we present a method to learn a bidirectional representation between the visual and muscle modalities. Section 5 shows experiments on both in-distribution experiments and subjects, as well as out-of-distribution experiments and subjects. In section 6, we showcase a demo application for learning the bidirectional representation between modalities. By releasing our datasets and models publicly, we hope this paper will spur additional work that models the rich internal structure that drives human activity in video.

\begin{figure*}[t]
\centering
%\framebox[4.0in]{$\;$}
\includegraphics[width=\linewidth]{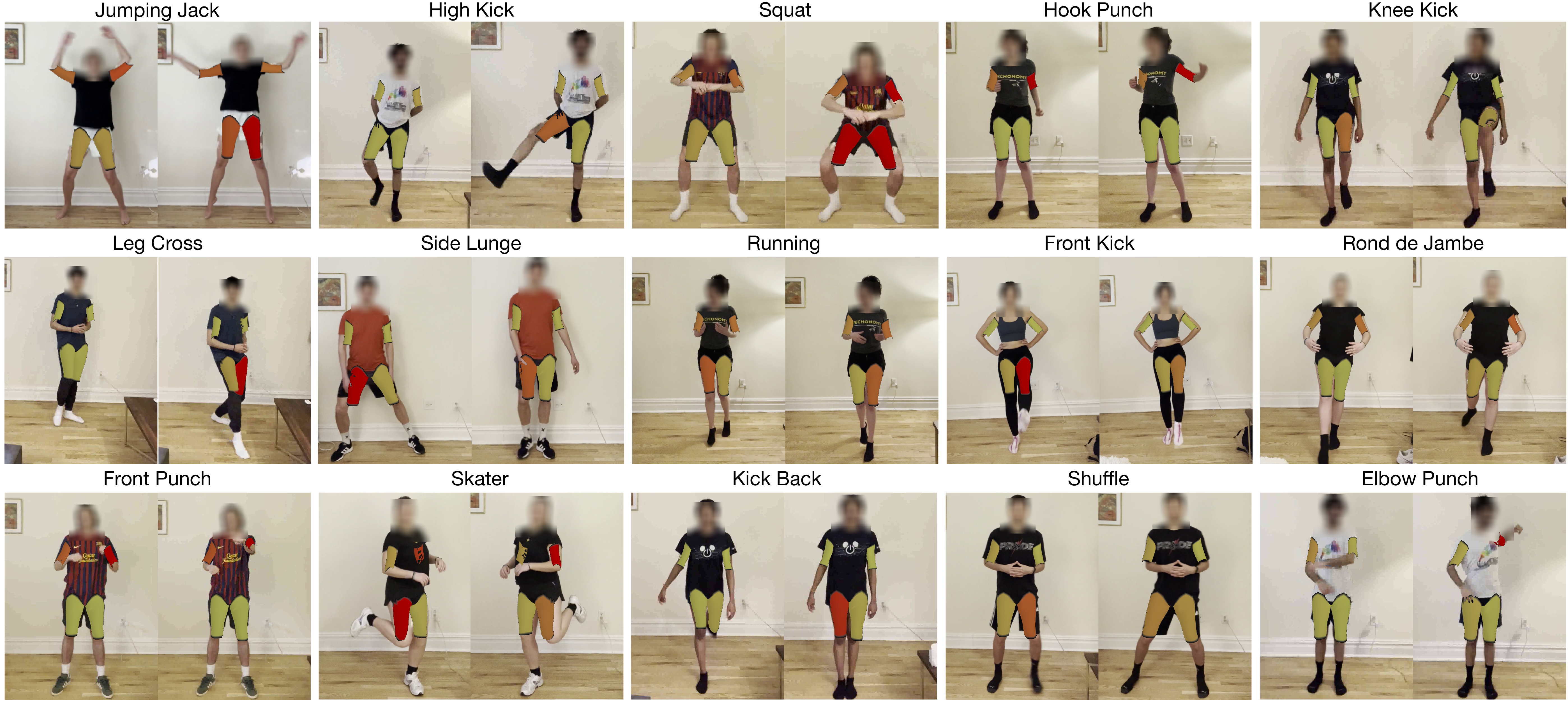}
\caption{\textbf{The MIA Dataset.} We illustrate two canonical frames from our dataset for each of the 15 exercises. }
 \label{fig:exercises}
\end{figure*}

\section{Related Work}

\textbf{Human Motion Prediction from Video.}
The field of computer vision has seen tremendous progress in inferring information about human motion from monocular video. One of the tasks is to regress human pose from video by regressing skeleton key-points and meshes \cite{loper2015smpl, kocabas2020vibe, yuan2022glamr, pishchulin2016deepcut, cao2017realtime, kay2017kinetics, cao2017realtime}. A related task is action segmentation \cite{lea2017temporal}. Other tasks that span from the pose estimation results include human motion transfer \cite{chan2019everybody, aberman2019deep, liu2019neural, wang2018video} and even pose correction to make a given pose anatomically-correct \cite{isaac2022single,rohan2020human}. Closely related, Park et al. predict the 3D gravity direction from a moving first-person view using inverse dynamics \cite{park2016force}.

% Additionally, action classification and recognition are examples of the vision community leveraging video to predict information about various human motions and behaviors \cite{raptis2013poselet,vahdat2011discriminative,kong2017deep,kong2014discriminative,wang2019progressive}, as well as action intentionality \cite{epstein2020oops}. Others have even tried to predict the future actions in videos \cite{abu2018will,vondrick2016anticipating,kitani2012activity,ryoo2011human,hoai2014max}. 

\textbf{Conditional Human Motion Generation.} The field of conditional human motion generation is well-established, with a diverse set of conditioning mechanisms. There are works that condition based on past frames and/or future target frame(s) \cite{fragkiadaki2015recurrent, martinez2017human, hernandez2019human, guo2023back, harvey2018recurrent,kaufmann2020convolutional,harvey2020robust, duan2021single, tevet2022human}. Others condition based on other modalities such as spatial trajectory \cite{holden2016deep}, action class \cite{guo2020action2motion,petrovich2021action,tevet2022human}, natural language text \cite{ahuja2019language2pose,petrovich2022temos,tevet2022human}, as well as audio \cite{li2021ai,aristidou2021rhythm}. In this work, our motion generation is conditioned on an input motion, as well as muscle activity constraints.

\looseness=-1\textbf{Multi-Modal Representations.}
 Multi-modal learning with video is a long-standing problem in computer vision. Some works predict other modalities from video, such as sound \cite{owens2016visually, gan2020foley} by using the natural temporal correspondence between video and sound, as well as video captioning \cite{xu2021videoclip}. Beyond prediction, multi-modal learning has been shown to improve video representations, for example from sound \cite{owens2016ambient,arandjelovic2017look}, and language \cite{xu2021videoclip,sun2019videobert}. The converse has also been explored - leveraging large datasets of unlabelled video has improved representations of other modalities  \cite{aytar2016soundnet,arandjelovic2017look, ma2020active,gao2020listen}. We are interested in predicting an entirely different modality from monocular video, muscle activation, as well as reconstructing motion from muscle activity.

\textbf{Electromyography.} To measure muscle activation, we use Surface Electromyography (sEMG) sensors, which are attached to electrodes placed on human skin to measure the electrical activity of muscle tissue. Reconstructing parts of human pose from sEMG data is an established task, but only for either arms \cite{liu2021neuropose,abraham2015arm,nasri2019inferring,quivira2018translating} specifically, or legs \cite{zhang2012semg} separately. In the forward direction, previous work has tackled predicting muscle activation, however, the input modality has not been video. Some works use torque or surface force measurements as input \cite{sekiya2019linear,song2015inverse,li2014inverse}, while others use goniometers to track pose \cite{tamilselvam2021musculoskeletal} or motion capture tracking systems \cite{johnson2009evaluation,yamane2009muscle,nakamura2005somatosensory}. Our work seeks to predict muscle activation with no additional hardware at test-time besides video. Additionally, certain works predict muscle activation directly from 3D point clouds collected by depth cameras \cite{niu2022estimating,sagawa2018predicting}. However, these works rely on seeing the skin deformation on the human subject to infer muscle activity. We infer muscle activity from motion priors, not the visible increase in muscle size. This allows our model to work for clothed humans, or humans exercising at a distance. 

\looseness=-1\textbf{Modeling Human Motion with Physics.} Recent work has focused on generating motion that respects the physics of motion via physics simulations of human motion dynamics \cite{yuan2022physdiff,luo2021dynamics,yuan2021simpoe}. However, the simulated humanoids are constructed with assumptions about the physics of human motion. Additionally, sEMG studies have shown that for a given motion, different people vary in how they recruit muscle groups to execute that motion \cite{trepman1998electromyographic}. Single humanoid simulations will not capture this diversity in real humans' motion dynamics.

%------------------------------------------------------------------------

\section{The Muscles in Action (MIA) Dataset}

To explore the mapping between visual motion and muscle activity, we collected a dataset of synchronized video and sEMG signals. Our dataset contains 15 different exercises, which each of the 10 subjects perform.

\begin{figure*}[t]
\centering
%\framebox[4.0in]{$\;$}
\includegraphics[width=\linewidth]{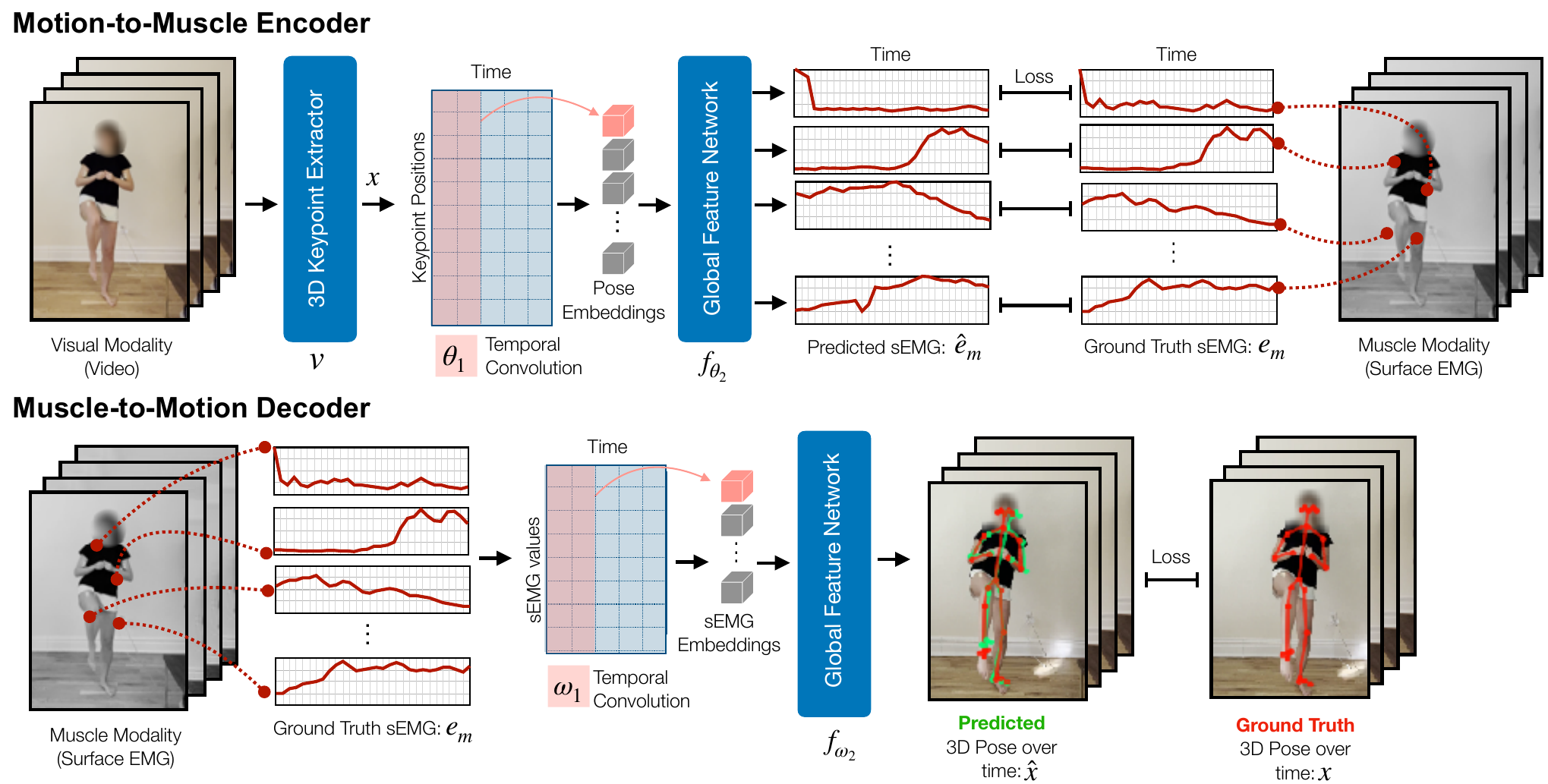}
\caption{\textbf{Encoder and Decoder Architectures.} We illustrate the architectures for our Motion-to-Muscle encoder and our Muscle-to-Motion decoder.}
\label{fig:arch}
\vspace{-1em}
\end{figure*}

\subsection{Data Collection}

Our dataset consists of 12.5 hours of synchronized video and sEMG signals, for eight muscles. These eight muscles include the left and right biceps brachii (biceps), the left and right latissimus dorsi (laterals), both quadriceps (quads), and both biceps femoris (hamstrings). The collected sEMG values correspond to the neuromuscular junction's total bioelectric energy. 

% We intentionally chose 4 types of kicks (front kick, back kick, high leg, knee kick), 3 types of punches (front punch, hook punch, elbow punch), 2 types of displacement exercises (running and shuffle),  2 type of side-to-side exercises (slow skater and leg cross), and four instances of unique (1) exercises (side lunge, jumping jack, rond de jambe and squat). This allows us to later compare how increased category prominence affects OOD results. Figure \ref{fig:exercises} illustrates two canonical frames for each exercise, with the ground-truth muscle activity illustrated for the frontal muscles: the biceps and the quadriceps.

\looseness=-1
The dataset consists of 15 exercises shown in Figure \ref{fig:exercises}. Each subject performed each exercise for 5 minutes, and we asked them to vary the execution's speed, effort, and orientation. There are a total of 10 subjects in the dataset, 5 of which are females and 5 of which are males. We collected 75 minutes of data for each subject, totalling 12.5 hours of data. The subjects varied in body weight and muscle. To collect the sEMG data, we used eight M40 Muscle Sense bluetooth wireless EMG sensors from ANR Corp. To collect the video, we used a standard iPhone 10 camera. Please see the supplementary for details on the electrode and sensor placement method.
% Following standard procedure (CITE), prior to placing the electrodes on the subjects' skin, we asked the subjects to shave the designated areas, which was followed by wiping the designated areas with alcohol, to optimize for clean signal capture. The sensors were placed on the specified muscles, and not moved during the entire data collection process. We did our best to position the sensors in the same location on the muscle across all subjects. 

%\subsection{Equipment}

\subsection{Data Preprocessing}

The sEMG data's sample rate is 10fps, and the each data point from the sensor comes with a timestamp. The iPhone video records at 29.97 fps, and also has a time-stamp. We resample the video to match the frame rate of the sEMG data, and use these time stamps to align the muscle and visual modalities. We explain our exact methodology for this in the supplemental material. 

Once the sEMG and the frames are aligned, we extract both 3D keypoints and 2D keypoints with the VIBE model and checkpoints  \cite{kocabas2020vibe}. The 3D keypoints are normalized with respect to a pre-computed bounding box, while the 2D keypoints are absolute with respect to the frame dimensions. For all experiments unless explicitly stated otherwise, the input sequence length is 30 frames and the output sequence length is 30 sEMG values per muscle, corresponding to 3 seconds. Once the dataset was split into intervals of 3 seconds, the train/test split was created by randomly choosing 20\% of the 3 second intervals within an exercise per subject to be allocated to the test set, and the remaining 80\% was allocated to the training set.

% There are two cases. In the first case, if the sEMG recording begins before the video, we simply remove all data prior to the video's origin time, and the first sEMG value that is larger or equal to the starting time gets rounded down to the nearest 10 millecond interval. The rest of the sEMG data is timestamped by adding 10ms to each data point. If the sEMG recording begins after the video, we remove the frames up until we reach a frame for which the timestamp is larger than the starting sEMG value. Then we repeat the processing steps for the first case.To summarize, the maximum misalignment between video and and sEMG data is less than 10ms. 

\section{Method}

Our approach aims to learn the bidirectional mapping between the visual modality and the muscle modality, which allows us to perform three tasks: a) infer muscle activity from video, b) infer pose from muscle activity, and c) provide recommended motions to people that will target certain muscles. In this section, we present this approach.

\subsection{Muscle and Motion Mappings}

\looseness=-1
The characteristics of muscle activity make the sEMG signal challenging to analytically process.  We aim to overcome these challenges by leveraging the synchronization with the visual modality. By finding the correlations between a person's visible motion and the sEMG signal, we can learn representations that encode muscle activity with respect to motion. 

% When a person moves, for the same movement, different directions of motion affect which muscles get engaged. For instance, during a squat, when a person moves downwards, the quads are not activated, as gravity does a good portion the heavy lifting. However, when the person rises back up, the quads are activated. Moreover, differences in direction can also change which muscles get activated. For example, a hook punch and a front punch differ only in the direction of the punch. However, a hook punch engages the lateral muscles significantly, while a front punch much less so. Finally, individuals vary in which muscles they engage. While certain muscles are necessary to execute the movement, each person will perform the motion slightly differently, and engage various muscle groups to different degrees.

Let $x \in \mathbb{R}^{KD \times T}$ be the human pose of a person, extracted over $K$ keypoints, with dimensionality $D$, for $T$ frames in a video. Our goal is to predict the muscle activity that created the motion, which we denote as $m \in \mathbb{R}^{M \times T}$ for $M$ individual muscles, as well as to reconstruct motion $x \in \mathbb{R}^{KD \times T}$ from muscle activity $m$. We aim to learn mappings that transform between these spaces through the functions: 
\begin{align}
    \hat{m} = E_\theta(x) \quad \textrm{and} \quad \hat{x} = D_\omega(m)
\end{align}
\looseness=-1 where $E_\theta(x)$ is an encoder parameterized by $\theta$ and $D$ is a decoder  parameterized by $\omega$, both of which are neural networks whose architecture we describe later. We learn the parameters for both models through the supervised learning problem:
\begin{align}
    \min_{\theta,\omega}\; \mathbb{E}_{(x,m)}\left[ \mathcal{L}\left(E_\theta(x), m\right) + \mathcal{L}\left(E_\omega(m), x\right)\right] 
\end{align}
where we use a mean squared loss function $\mathcal{L}$ to compare predictions to the ground truth in both modalities. We optimize both using stochastic gradient descent with the Adam optimizer \cite{kingma2014adam}. Full implementation details are provided in the supplemental material.

\subsection{Modification in Muscle Space}

% \textbf{Talk about how we do this. Don't need to repeat anything. We just need to explain the equation to multiply $m$ by a scalar, and explain the intuition for why this is correct and makes sense.}

In this section, we explain how our bidirectional model can be used to generate new motions based on the edits in the muscle modality. Given a goal to minimize a use of the muscle, or increase the workout of a muscle, we generate a new motion, similar to the input motion, with a modification that adheres to the muscle activity goal. To do so, given a video, our encoder $E$ first predicts muscle activation $\hat{m} \in \mathbb{R}^{M \times T}$, composed of $M$ sequences. Let $\hat{m}^{k} \in \mathbb{R}^{T}$ be one muscle sequences in particular that we choose to scale, either up or down, with scalar $s \in \mathbb{R}$:
\begin{align}
\bar{m}^{k} = s \cdot \hat{m}^{k} \quad \textrm{and} \quad \bar{m}^{j} = \hat{m}^{j} \; \forall_{j \ne k}
\end{align}
The new matrix $\bar{m} \in \mathbb{R}^{M \times T}$ is the edited $\hat{m}$ matrix. Our decoder $D$ decodes $\bar{m}$ into a recommended motion $\bar{x}$:
\begin{align}
\bar{x} = D(\bar{m})
\end{align}
This recommended motion $\bar{x}$ will be similar to the predicted reconstruction $\hat{x}$, except the recommended motion is in agreement with the muscle goals dictated by the edited predicted muscle activation $\bar{m}$.

 \begin{figure*}[t!]
 \centering
 \includegraphics[width=\linewidth]{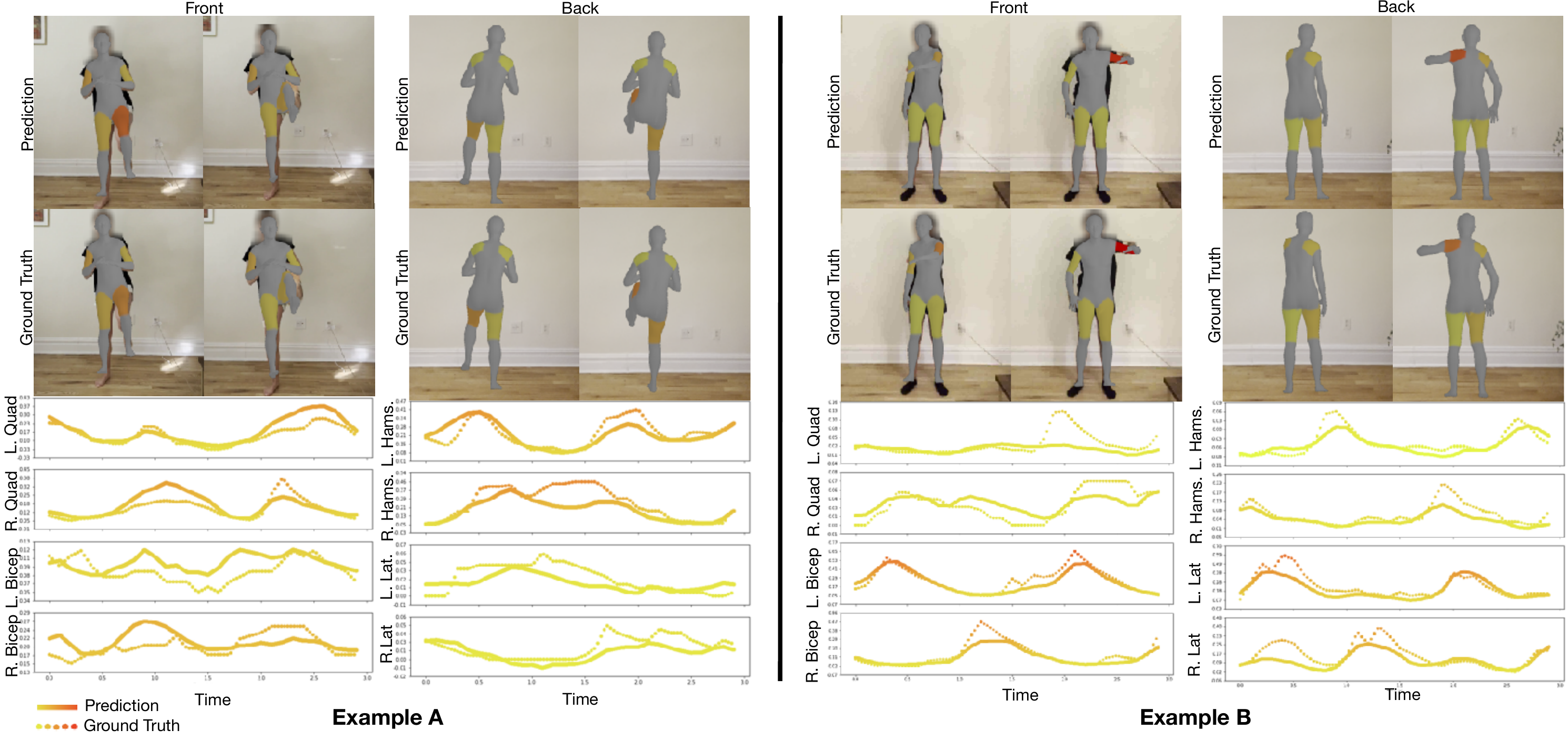}
 \caption{\textbf{Motion-to-Muscle Qualitative Results.} We illustrate two separate qualitative results. The first row of frames corresponds to a visualization of the predicted activations, and the second row of frames corresponds to a visualization of the ground truth activations. For the plot beneath the frames, the dotted line corresponds to the ground-truth values, and the solid line corresponds to the predictions. Yellow corresponds to relaxed, and red corresponds to flexed.}
 \vspace{-1em}
  \label{fig:qualresults}
 \end{figure*}

 \begin{figure*}[t]
\centering
%\framebox[4.0in]{$\;$}
\includegraphics[width=\linewidth]{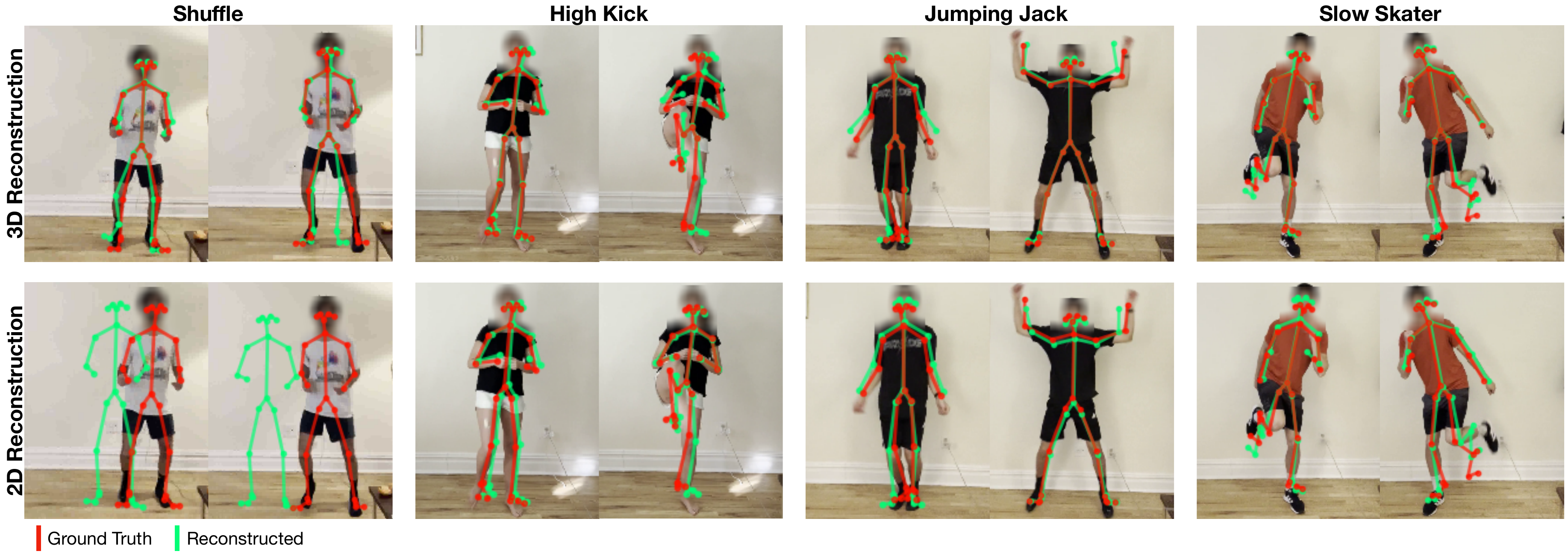}
\caption{\textbf{Muscle-to-Motion Qualitative Results.} We show reconstructed 3D pose and 2D pose from muscle activity, where 3D pose is normalized with respect to a bounding box, while the 2D pose is absolute with respect to the image frame. Since sEMG signals don't contain location information, the 2D model cannot reconstruct the subject in the right location.}
\vspace{-1em}
 \label{fig:exercises}

\end{figure*}

\subsection{Architectures}

\looseness=-1
We use a common architecture for both the encoder and decoder, with only minimal modifications between them to adapt to their input and output modalities. See Figure \ref{fig:arch} for an overview of both architectures. We factorize the architectures:
\begin{align}
E(x) = f_{\theta_{2}}\left( g_{\theta_{1}}\left(x\right) \right) \; \textrm{and} \; D(m) = f_{\omega_{2}}\left( g_{\omega_{1}}\left(m\right) \right)
\end{align}
where $g$ is a local feature extractor and $f$ is a global feature network.
 
The local feature extractor for the Motion-to-Muscle encoder receives keypoints $x \in \mathbb{R}^{KD \times T}$, where $K$ is the number of keypoints, $D$ is their dimensionality, and $T$ is the number of frames. This matrix is then convolved with a filter $\theta$ that has $c$ channels:
\begin{align}
g_\theta(x) = \theta \ast x 
\end{align}
The spatial dimension of the kernels $\theta$ spans the entirety of the key-point dimension, and the temporal dimension spans roughly a second of time. Each kernel outputs a feature $n$, where $n \in \mathbb{R}^{1 \times T}$. Since there are $c$ channels, the resulting output from the temporal convolution layer is a sequence of $T$ embeddings $d_{1}, ..., d_{T}$, s.t. $d_t \in  \mathbb{R}^{128}$. 

For the Muscle-to-Motion decoder, the local feature extractor has the same structure, except the input is the muscle activation $m \in  \mathbb{R}^{M \times T}$. The convolutional layer's spatial dimension is changed to span the entirety of the muscle dimension, $M$, and the temporal dimension still spans roughly a second of time. The output dimensionality is thus also a sequence of $T$ embeddings $d_{1}, ..., d_{T}$, s.t. $d_{t} \in  \mathbb{R}^{128}$. 

% Convolutions excel at detecting subtle motion patterns, which makes this a very good embedding for the next part of the architecture. Additionally, convolution is designed for spatial invariance which is why we chose to make the spatial dimension of the kernel match the number of coordinates, otherwise motion patterns for legs would need to correspond to similar muscle activation patterns as motion patterns for arms, for example. However, the same motion in pattern in different limbs correspond to drastic differences in muscle activations, as previously explained by comparing gravity's relationship to hands and feet. The local motion encoder is local temporally, but global spatially.

The second part of our common architecture, the global feature network, needs to identify long range patterns over time in order to capture the dynamics of the sequence. It is global temporally. We implement it using a Transformer \cite{vaswani2017attention} with 4 layers with 8 attention heads  and no attention masking. The input to the Transformer is the sequence of embeddings $d_{1}, ..., d_{T}$, s.t. $d_{t} \in  \mathbb{R}^{128}$. The output of the Transformer is the sequence of embeddings $o_{1}, ..., o_{T}$, s.t. $o_{t} \in  \mathbb{R}^{128}$. For the Motion-to-Muscle encoder, a fully connected layer maps $o_{t} \in \mathbb{R}^{128}$ to a sequence of embeddings $m_{1},...,m_{T}$ s.t. $m_{t} \in \mathbb{R}^{M}$. For the Motion-to-Muscle decoder, a fully connected layer maps $o_{t} \in \mathbb{R}^{128}$  to $m_{t} \in \mathbb{R}^{KD}$.

\subsection{Conditioning}
For similar motions, different people will vary in muscle activities. This is mostly a product of three factors: a) slight variation in the motion itself b) different muscle recruitment due to personal style \cite{trepman1998electromyographic} c) slight variations in sensor placement and differences in morphology \cite{hermens2000development}. As such, we construct two additional conditional versions of our encoder and decoder. To accommodate the conditioning, we concatenate a unique tensor $y \in \mathbb{R}^{2 \times T}$ to the sequence of embeddings $d_{1},..,d_{T}$, per subject. Further details can be found in the supplementary. 

\section{Experiments}

The objective of our experiments is to analyze the alignment between the visual modality and the muscle activities underlying motion. We show results across the 15 exercises, using root mean squared error as our metric for both tasks.
\subsection{Baselines}

\textbf{Retrieval (Retr.).}
Our first baseline for solving this problem is to perform nearest neighbor. For the Motion-to-Muscle task, given an example 3D skeleton over time $x$ in the test set $X_{test}$, we retrieve the nearest neighbor $\bar{x}$ from $X_{train}$, and assign $\bar{x}$'s muscle activation $\bar{m}$ to $\hat{m}$ as the predicted muscle activation. For the Muscle-to-Motion task, given an example sequence of muscle activity $m$ in the test set $M_{test}$, we retrieve the nearest neighbor $\bar{m}$ from $M_{train}$, and assign $\bar{m}$'s 3D skeleton over time $\bar{x}$ to $\hat{x}$.

\textbf{Conditional Retrieval (C-Retr.).} Our second baseline is conditional retrieval, where we condition on the subject. For the Motion-to-Muscle task, given an example sequence of 3D pose $x$ for subject $s$, in the test set $X_{test}^{s}$, we retrieve the nearest neighbor $\bar{x}$ from $X_{train}^{s}$, which only contains data from subject $s$, and assign $\bar{x}$'s muscle activation sequence $\bar{m}$ to $\hat{m}$ as the predicted muscle activation. We report the average across subjects. The same method is applied to the Muscle-to-Motion task.

\begin{table}[b]
\resizebox{\columnwidth}{!}{%
\setlength{\tabcolsep}{2pt}
\centering
  \begin{tabular}{l|rrrr|rrrr}
    \toprule
    & \multicolumn{4}{c|}{In-Distribution Encoder} & \multicolumn{4}{|c}{Out-of-Distribution Encoder} \\
      Exercise & Retr. & C-Retr. & Ours & C-Ours & Retr. & C-Retr. & Ours & C-Ours\\
    \midrule
    ElbowPunch & 15.1 & 15.2 & 12.0 & \textbf{12.0} & 25.7 & 29.4 & 19.8 & \textbf{19.7}\\
    FrontKick & 10.5 & 9.8 & \textbf{7.8} & 7.9 & 32.7 & 54.4 & 11.0 & \textbf{11.0}\\
    FrontPunch & 10.9 & 10.7 & 8.7 & \textbf{8.6} & 27.6 & 22.8 & 15.9 & \textbf{15.5}\\
    HighKick & 13.0 & 12.8 & 10.1 & \textbf{10.1} & 17.6 & 17.8 & 15.8 & \textbf{15.5}\\
    HookPunch & 16.4 & 16.3 & 12.5 & \textbf{12.4} & 23.4 & 23.6 & 19.3 & \textbf{18.9}\\ 
    JumpingJack  & 25.7 & 25.4  &  19.2 & \textbf{19.2} & 47.6 & 47.0 & \textbf{37.0} & 41.0\\
    KneeKick  & 11.2 & 10.7 & 8.2 & \textbf{8.0} & 16.7 & 15.5 & 13.2 & \textbf{12.8}\\
    KickBack & 12.3 & 11.7 & 9.3 & \textbf{9.3} & 17.5 & 19.2 & \textbf{15.3} & 15.6\\
    LegCross & 12.0 & 10.2 & 8.0 & \textbf{8.0} & 18.1 & 16.7 & \textbf{15.2} & 15.4 \\
    RonddeJambe & 23.8 & 23.7 & 20.4 & \textbf{20.3} & 36.8 & 35.0 & 33.4 & \textbf{33.2} \\
    Running & 15.8 & 10.6 & 8.7 & \textbf{8.6} & 27.2 & 15.6 & 14.0 & \textbf{14.0}\\ 
    Shuffle & 13.6 & 13.2 & 9.9 & \textbf{9.9} & 22.1 & 17.0 & 14.2 & \textbf{14.0}\\
    SideLunge & 17.2 & 16.5 & 13.8 & \textbf{13.7} & 27.7 & 30.1 & 24.4 & \textbf{24.0}\\
    SlowSkater & 16.8 & 16.3 & 13.1  & \textbf{11.4} & 25.2 & 23.0 & 21.1 & \textbf{20.8}\\
    Squat  & 20.2 & 19.8 & \textbf{15.9} & 16.0 & 36.3 & 34.0 & 35.2 & 30.4\\
    \bottomrule
  \end{tabular}}
    \caption{\textbf{RMSE per Exercise for the Encoder}. We report the rMSE per exercise 
 for muscle prediction.}
    
  \label{table-one}
\end{table}

 \begin{table}[t]
\resizebox{\columnwidth}{!}{%
\setlength{\tabcolsep}{2pt}
\centering
  \begin{tabular}{l|rrrr|rrrr}
    \toprule
    & \multicolumn{4}{c|}{In-Distribution Decoder} & \multicolumn{4}{|c}{Out-of-Distribution Decoder} \\
      Exercise & Retr. & C-Retr. & Ours & C-Ours & Retr. & C-Retr. & Ours & C-Ours\\
    \midrule
    ElbowPunch & 0.045 & 0.43 & 0.031 & \textbf{0.031} & 0.078 & 0.078 & 0.060 & \textbf{0.060}\\
    FrontKick & 0.058 & 0.052 & \textbf{0.040} & 0.043 & 0.103 & 0.099 & \textbf{0.074} & 0.077\\
    FrontPunch & 0.047 & 0.045 & \textbf{0.032} & 0.033 & 0.075 & 0.076 & 0.062 & \textbf{0.061}\\
    HighKick & 0.093 & 0.090 & 0.076 & \textbf{0.074} & 0.145 & 0.139 & 0.119 & \textbf{0.119}\\
    HookPunch & 0.060  & 0.055 & \textbf{0.044} &0.045 & 0.090 & 0.087 & \textbf{0.075} & 0.076\\ 
    JumpingJack & 0.071 & 0.07 & \textbf{0.062}  & 0.066 & 0.143 & 0.146 & 0.109 & \textbf{0.108}\\
    KneeKick  & 0.079 & 0.077 & \textbf{0.061} & 0.064 & 0.119 & 0.114 & 0.096 & \textbf{0.096}\\
    KickBack & 0.086 & 0.082 & 0.072 & \textbf{0.069} & 0.116 & 0.114 & 0.093 & \textbf{0.093}\\
    LegCross & 0.056 & 0.049 & \textbf{0.040} & 0.042 & 0.112 & 0.106 & 0.087 & \textbf{0.087} \\
    RonddeJambe & 0.074 & 0.069 & \textbf{0.055} & 0.056 & 0.119 & 0.116 & \textbf{0.092} & 0.093 \\
    Running & 0.047 & 0.046 & 0.037 & \textbf{0.037} & 0.074 & 0.070 & 0.052 & \textbf{0.051}\\ 
    Shuffle & 0.058 & 0.056 & \textbf{0.043} & 0.044 & 0.077 & 0.072 & \textbf{0.056} & 0.057\\
    SideLunge & 0.07 & 0.067 & 0.058 & \textbf{0.057} & 0.127 & 0.123 & 0.108 & \textbf{0.108}\\
    SlowSkater & 0.076 & 0.072 & \textbf{0.063}  & 0.065 & 0.140 & 0.124 & 0.109  & \textbf{0.109}\\
    Squat  & 0.066 & 0.064 & \textbf{0.057}  & 0.059 & 0.132 & 0.126 & \textbf{0.111}  & 0.112 \\
    \bottomrule
  \end{tabular}}
    \caption{\textbf{RMSE per Exercise for the Decoder.} We report the rMSE per exercise 
 for motion prediction.}
    
  \label{table-two}
\end{table}

\begin{table}[t]
\resizebox{\columnwidth}{!}{%
\setlength{\tabcolsep}{5pt}
\centering
  \begin{tabular}{l|ccccccc}
    \toprule
      Frame Count & 1 & 5 & 10 & 15 & 20 & 25 & 30\\
    \midrule
    C-Retr.  & 19.6 & 16.3 & 14.5 & 14.0 & 13.5 & 13.3 & 14.3\\
    C-Ours & 13.3 & 11.1 & 10.3  & 9.6  & 9.1 & 8.8 & 8.8\\
    \bottomrule
  \end{tabular}}
    \caption{\textbf{Temporal Analysis}. We report the root mean squared error for the conditional baseline as well as for our conditional model as we change the length of the sequences.}
    \vspace{-1em}
  \label{table-three}
\end{table}

 % \begin{figure}[t!]
 % \centering
 % \includegraphics[width=\linewidth]{figures/rmse_time.pdf}
 % \caption{\textbf{Temporal analysis.} We plot the performance of our conditional baseline and our conditional model with respect to the number of frames the model receives and predicts.}
 %  \label{fig:editsquals}
 % \end{figure}

\subsection{Results}

In this section, we report quantitative and qualitative results for both the encoder and the decoder. For the quantitative results, we report the results per exercise. We report these results for our conditional and non-conditional baselines and models.

\textbf{Motion-to-Muscle Encoder.} As seen in Table \ref{table-one}, for all of the 15 exercises, both our conditional and non-conditional model outperform both conditional and non-conditional retrieval baselines. For the out-of-distribution experiments, we retrained 15 models on 15 different datasets, each dataset leaving out one exercise. For each model, we then ran inference on the exercise that was left out and reported it in the last four columns of Table \ref{table-one}, denoted as Out-of-Distribution Encoder. The results indicate that our learning method generalizes better to unseen exercises than the retrieval baselines. Finally, in both the in-distribution and out-of-distribution experiments, and in both our model and baselines, we observe that for most exercises, the conditional model outperforms the non-conditional model. This result confirms the hypothesis presented in Section 4.4.  

We show two qualitative examples of predicting muscle activation from motion in Figure \ref{fig:qualresults}. The first column shows a subject performing a high kick, and the second column shows a subject performing an elbow punch. The axis in the third and fourth rows are scaled to the range of values per example per muscle, and we denote the absolute value with a gradient from yellow (low activation) to red (high activation). Even when the range of the sEMG signal is small, indicated by a plot that stays mostly one color, we notice that the predictions follow the ground-truth fairly closely. The alignment of the prediction and ground-truth is often close, showing a small error in phase. We show more qualitative results in the supplementary material. 

\textbf{Temporal Analysis.} In order to evaluate how crucial the temporal component
of our model was, we retrained seperate 7 separate transformer models to predict muscle activation from motion, where both the input and output are $n$ frames for 7 different input/output lengths. We notice that for both the conditional baseline and our conditional model, the performance increases as the model sees examples with longer temporal length. However, the conditional baseline drops in performance from 25 to 30 frames, whereas our conditional model does not. 

 \begin{figure*}[t!]
 \centering
 \includegraphics[width=\linewidth]{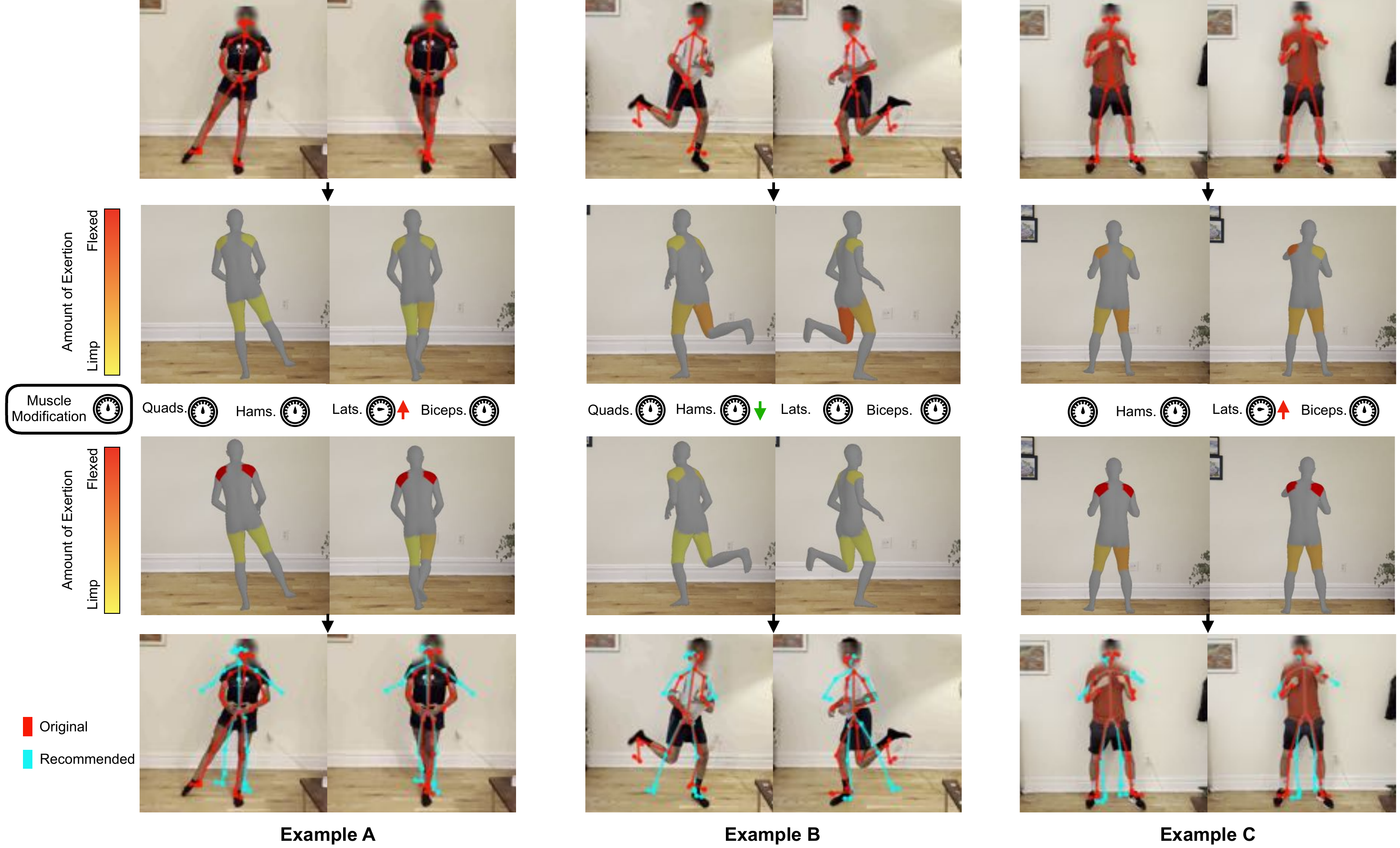}
 \caption{\textbf{Qualitative Results.} We illustrate three more qualitative results for the editing task. The first row illustrates the ground truth 3D skeleton projected onto the frame. The second row shows the predicted muscle activation for the dorsal muscles (laterals and hamstrings). Since we are visualizing the back of the person, the meshes are flipped. The third row shows the scaled predicted muscle activations. The fourth row illustrates the recommended motion which the decoder generates from the scaled predicted muscle activations.}
 \vspace{-1em} 
  \label{fig:editsquals}
 \end{figure*}

\textbf{Muscle-to-Motion Decoder.} Similarly, we report the root mean squared error per exercise, for both the conditional decoder and non-conditional decoder in Table \ref{table-two}. For all of the 15 exercises, both our conditional and non-conditional model outperforms both conditional and non-conditional retrieval baselines. For the decoder, there is less of a clear pattern between the conditioned model and the non-conditioned model. We believe that this is explained by the fact that muscle activity already has conditioning embedded within it. Subjects often have trademark muscles, that they use more or less, or with different ranges. As such, explicit conditioning may not be helpful. 

We also show four qualitative examples from our Muscle-to-Motion decoder in Figure \ref{fig:exercises}. The first row illustrates results from our main decoder, which regresses to 3D pose over time. The second row illustrates a secondary decoder, which regresses to 2D pose over time. The extracted 3D keypoints from VIBE \cite{kocabas2020vibe} are normalized with respect to a given bounding box, which we utilize to project the predicted 3D keypoints onto the 2D image. This is why our 3D pose decoder results have the subject in the right location, even for exercises that have high displacement, such as shuffle. On the contrary, the 2D Decoder, whose ground-truth coordinates are not normalized with respect to a bounding box, is unable to predict the subject's displacement since muscle activation has no information about a subject's location.  We show more qualitative results in the supplementary material.

\section{Editing}

We show examples of how the motion-muscle mappings can be leveraged to generate motion recommendations subject to muscle constraints in Figures \ref{fig:teaser} and  \ref{fig:editsquals}. Given a motion, we predict the muscle activation, which we edit one or more muscle predictions, as described in Equations 2 and 3 in Section 4, and decode the edited predicted muscle activation into a recommended motion. 

When the modification amplifies the muscle, then it generates a corresponding motion with minimal change that only causes the exercise to engage the target muscle more. For example, in Figure \ref{fig:editsquals}A, shows a person performing the Rond de Jambe, however their use of laterals is low. By amplifying the muscular representation, the generated motion lifts the arms up correctly, therefore engaging the laterals more. We see a similar trend in Figure \ref{fig:editsquals}C. 

There is a converse effect when the modification attenuates the muscle. Figure \ref{fig:editsquals}B shows a person performing a slow skater, which due to the non-supporting leg bending backwards, activates the hamstrings significantly. By modifying the muscular representation to attenuate the hamstrings, the generated motion prevents the non-supporting leg from bending backwards, disengaging the hamstrings.

Moreover, the temporal pattern of the recommended motion matches that of the input motion, as the only edit performed is scaling. This is useful for AR/VR applications. 

% \textbf{Increasing Engagement} One use-case for this application is to increase muscle engagement. In the first example in Figure \ref{fig:teaser}, we increase the activation of the quads, hamstrings and laterals. The generated motion is a squat with legs are further apart, reflecting more muscle engagement. In Figure \ref{fig:editsquals}, for examples A and C, we increase predicted muscle activation for the laterals. In both cases, the arms are lifted. While the dataset includes elbow punches, the direction of an elbow punch is opposite to the direction of a front punch. However, the generated motion in example C maintains the direction of motion of the front punch, yet learns to raise the starting point of the front punch higher. This demonstrates our model's ability to compose motions.

 % A second use-case for this application would be to decrease a particular muscle group engagement, for instance if the subject has injured the particular muscle. Example B illustrates a subject performing a slow skater. The predicted muscle activity is high for hamstrings. We decrease the predicted muscle activity for the hamstrings and the decoded recommended motion prevents the non-supporting leg from bending backwards, thus disengaging the hamstrings.

\section{Conclusion}
This paper presents a new multi-modal dataset, the Muscles in Action (MIA) dataset, for modeling the relationship between muscle activity and motion. We present our framework for learning the bidirectional mapping between the modalities. We also demo how our bidirectional model can be used to generate recommended motions conditioned on muscle activity objectives.
%-------------------------------------------------------------------------

\textbf{Acknowledgements:} We would like to thank our subjects for participating in the dataset. We'd also like to thank Jianbo Shi, Georgia Gkioxari, Huy Ha and Kamyar Ghasemipour for their helpful feedback. This research is based on work partially supported by the NSF NRI Award \#1925157. M.C.\ is supported by the Amazon CAIT PhD fellowship. The views and conclusions contained herein are those of the authors and should not be interpreted as necessarily representing the official policies, either expressed or implied, of the sponsors.

{\small
\bibliographystyle{ieee_fullname}
\bibliography{egbib}
}

\end{document}